\documentclass{article}

     \PassOptionsToPackage{numbers, compress}{natbib}

     \usepackage[final]{neurips_2019}

\usepackage[utf8]{inputenc} 
\usepackage[T1]{fontenc}    
\usepackage{hyperref}       
\usepackage{url}            
\usepackage{booktabs}       
\usepackage{amsfonts}       
\usepackage{nicefrac}       
\usepackage{microtype}      
\usepackage{float}
\usepackage{graphicx}
\usepackage{subcaption}

\title{Few-shot tweet detection in emerging disaster events}

\author{%
  Anna Kruspe \\
  German Aerospace Center (DLR)\\
  Institute of Data Science\\
  Jena, Germany \\
  \texttt{anna.kruspe@dlr.de} \\
}

\begin{document}

\maketitle

\begin{abstract}
Social media sources can provide crucial information in crisis situations, but discovering relevant messages is not trivial. Methods have so far focused on universal detection models for all kinds of crises or for certain crisis types (e.g. floods). Event-specific models could implement a more focused search area, but collecting data and training new models for a crisis that is already in progress is costly and may take too much time for a prompt response.
As a compromise, manually collecting a small amount of example messages is feasible. Few-shot models can generalize to unseen classes with such a small handful of examples, and do not need be trained anew for each event. We compare how few-shot approaches (matching networks and prototypical networks) perform for this task. Since this is essentially a one-class problem, we also demonstrate how a modified one-class version of prototypical models can be used for this application.
\end{abstract}

\section{Introduction}\label{sec:intro}
Social media is an interesting source of information during disasters that has become a topic of research in recent years. Twitter users, as an example, write about disaster preparations, developments, recovery, and a host of other topics \cite{niles}. Retrieving this information could lead to significant improvements in disaster management strategies. According to a Red Cross study, $69\%$ of Americans think that emergency response agencies should respond to calls for help sent through social media channels \cite{redcross}. The crux of this matter lies in the retrieval and classification of such messages. Twitter users generate 5,800 tweets per second on average\footnote{\url{https://www.omnicoreagency.com/twitter-statistics/}}. Even with hashtag- or location-based pre-filtering, sophisticated automatic methods are necessary to detect disaster-related messages in acceptable timespans. Approaches published so far have mainly had the generalized detection of any disaster-related tweet as their objective \cite{burel}, or have focused on specific types of disasters \cite{caragea}. Designing models specific to a certain event could potentially yield much more exact results, but training such dedicated models would require large amounts of already available data and take up critical time in a disaster.\\

One- or few-shot models offer the possibility to detect instances of an unseen class on the basis of a small number of example instances. In this paper, we demonstrate how to use such models to detect Twitter messages (tweets) pertaining to a specific event on the basis of example tweets. These models can be trained well in advance, and then require as few as ten examples against which new tweets are compared. We test few-shot networks that either require both positive and negative examples of the event, or only require positive examples. 

\section{Related work}\label{sec:sota}
As described above, users generate huge amounts of data on Twitter every second, and finding tweets related to an ongoing event is not trivial \cite{landwehr}. Several detection approaches have been presented in literature so far.

The most obvious strategy is the filtering of tweets by various surface characteristics as shown in \cite{kumar}, for example. Keywords and hashtags are used most frequently for this and often serve as a useful pre-filter. Olteanu et al. developed a lexicon called \textit{CrisisLex} for this purpose \cite{olteanu}. However, this approach easily misses tweets that do not mention the keywords specified in advance, particularly when changes occur or the attention focus shifts during the event. It may also retrieve unrelated data that contains the same keywords \cite{imran2}. Geo-location is another frequently employed feature that can be useful for retrieving tweets from an area affected by a disaster. However, this approach misses important information that could be coming from a source outside the area, such as help providers or news sources. Additionally, only a small fraction of tweets is geo-tagged at all, leading to a large amount of missed tweets from the area \cite{sloan}. To resolve these problems, several other strategies were developed, starting with crowdsourcing platforms. On these platforms, a large amount of users hand-selects and labels incoming tweets in disaster situations. Examples include \textit{Ushahidi}\footnote{\url{https://www.ushahidi.com/}} and \textit{CrisisTracker} \cite{rogstadius}. Some of them already integrate ``traditional'' machine learning models, e.g. \textit{AIDR} \cite{imran2}.

In recent years, approaches based on deep learning techniques have come to the forefront of research. On a more general level, the problem falls under the umbrella of event detection as shown, for example, in \cite{chen,feng,nguyen2}. Caragea et al. first employed Convolutional Neural Networks (CNN) for the classification of tweets into those related to flood events and those unrelated \cite{caragea}. In many of the following approaches, a type of CNN developed by Kim for text classification is used \cite{kim}, such as in \cite{burel}. This method achieves an accuracy of $80\%$ for the classification into related and unrelated tweets. In the same publication as well as in \cite{burel2} and \cite{nguyen}, this kind of model is also used for information type classification. For comparison purposes, we also tested the relatedness model from \cite{burel} on our test data set (\textit{CrisisNLP}, see below) and obtained an $F_1$ measure of $.86$ and an accuracy of $.77$, although the data in this case is strongly unbalanced as there are few tweets in that data set that do not belong to any crisis.

All of these approaches aim to generalize to any kind of event without any \textit{a priori} information. A real-world system may not need to be restricted in this way; in many cases, its users will already have some information about the event, and may already have spotted tweets of the required type. This removes the need to anticipate any type of event. It also directs the system towards a specific event rather than any event happening at that time. Alam et al. \cite{alam} propose an interesting solution to this: Their system includes an adversarial component which can be used to adapt a model trained on a specific event to a new one (i.e. a new domain).

We propose a system that does not assume an explicit notion of relatedness vs. unrelatedness (or relevance vs. irrelevance) to a crisis event. These qualities are not easy to define, and might vary for different users or different types of events. Additionally, as in \cite{alam}, we are interested in a method that is specific to an event rather than attempting to detect any kind of crisis-related tweet. In this paper, we demonstrate how to implement a system that is able to determine whether a tweet belongs to a class (i.e. crisis event) implicitly defined by a small selection of example tweets. We employ few-shot models to for this purpose; an overview over work related to this topic is given later.

\section{Data}\label{sec:data}
\subsection{Data sets}

We employ four tweet data sets to train and test our models: \textit{CrisisLexT26} and \textit{CrisisNLP} are two  widely-used collections of disaster-related tweets, while \textit{Events2012} and \textit{Sentiment140} contain tweets spanning a wide range of topics.
\begin{description}
 \item[CrisisLexT26] \textit{CrisisLex} was first published by Olteanu et al. in 2014 \cite{olteanu} and expanded later to \textit{CrisisLexT26} \cite{olteanu2}. It contains tweets collected during 26 crises, mainly natural disasters like earthquakes, wildfires and floods, but also human-induced disasters like shootings and a train crash. Amounts of these tweets per disaster range between 1,100 and 157,500. In total, around 285,000 tweets were collected. They were then annotated by paid workers on the \textit{CrowdFlower} crowdsourcing platform\footnote{Now named \textit{Figure Eight}, \url{https://www.figure-eight.com/}} according to three concepts: Informativeness, information type, and tweet source. In this work, a balanced set containing 1,100 English-language tweets per event is used. Two events were excluded as no frequent hashtags for them were found.
\item[CrisisNLP] Similar to \textit{CrisisLexT26}, the team behind \textit{CrisisNLP} collected tweets during 19 natural and health-related disasters and published them for research \cite{imran}. Collected tweets range between 17,000 and 28 million per event, making up around 53 million in total. Out of these, around 50,000 were annotated both by volunteers and by paid workers on \textit{CrowdFlower} with regard to information type. In this work, only the tweets with \textit{CrowdFlower} annotations were utilized. These tweets come from a subset of 11 English-language event sets, summing up to around 23,000 in total. There is no overlap between the events in both data sets.
\item[Events2012] This data set contains 120 million tweets, of which around 150,000 were labeled to belong to one of 506 events (which are not necessarily disaster events) \cite{mcminn}. We use a subset of 90,000 of these tweets as not all original ones are still available. There is no overlap between the events in both previously mentioned data sets and this one.
\item[Sentiment140] For a wider selection of non-disaster tweets, we addnother data set called \textit{Sentiment140} \cite{go}. \textit{Sentiment140} was originally developed for sentiment detection in tweets, but these annotations are not used in this work. The whole data set contains 1.6 million tweets; to reduce processing times, we only use a random subset of 50,000.
\end{description}

\subsection{Experimental data composition}\label{subsec:data_comp}
In order to train and test our few-shot models, sub-sets of positive and negative supports for a class (examples and counterexamples) plus an either positive or negative query are necessary. Training steps performed on such sets are called ``episodes''. For our purposes, tweets are considered to belong to the same class if they are associated with the same event. We perform experiments on two different constellations of data: Event-vs.-event training, and event-vs.-all training.
\begin{description}
 \item[Event-vs.-event] In these experiments, we attempt to detect tweets related to a specific disaster event out of tweets related to other disaster events.\textit{CrisisLexT26} is used to generate the training episodes, and \textit{CrisisNLP} is used for the validation episodes. To generate episode data packs, tweets for an event containing a specific hashtag are selected randomly to create the positive support set. Upper- and lower-case versions are equalized, and this hashtag is removed in a pre-processing step so as not to bias the classifier towards it. For positive queries, tweets without the hashtag, but coming from the same event are picked randomly. For negative supports and queries, random tweets from other events are used. Hashtags were chosen manually according to their prevalence in the event's tweets. An overview is given in tables \ref{tab:hashtags_crisislex} and \ref{tab:hashtags_crisisnlp}. The practical reasoning for this is that during an ongoing event, a user could quickly search for example tweets by a certain hashtag, then use the found examples to detect more event-related tweets without this hashtag using the models proposed by us. (Of course, a user may just as well use any other method for selecting examples of the class they are interested in).
 \item[Event-vs.-all] For the second set of experiments, we train models for the more general task of detecting tweets pertaining to a specific event out of random other tweets. Episodes are generated similarly to event-vs.-event training, but selection of positive and negative examples is based on the event set these tweets come from rather than contained hashtags. Training episodes are generated from \textit{Events2012}, and validation episodes from the combined \textit{CrisisLexT26} and \textit{Sentiment140} data sets. The \textit{Sentiment140} data is used only for generating negative support and query examples.
\end{description}

\newpage
\section{Models}
\vspace{-7pt}
\textbf{Matching networks}, which implement the one-shot learning idea, were first introduced by Vinyals et al. in 2016 \cite{vinyals}. Training is performed on sets of \textit{support} examples for a subset of possible classes plus one or more \textit{query} examples belonging to one of the classes in so-called \textit{episodes}. These episodes are generated for a wide range of class permutations. In the few-shot case, multiple support examples per class are treated independently of each other throughout the network. Example-related likelihoods are then summed into class likelihoods at the output. In this way, matching networks essentially implement a weighted nearest-neigbor classification. 
Matching networks contain two input branches: One for the support examples, and one for the query example. Both branches perform an embedding of the input; their weights may be shared. The resulting embedding of a query is then compared against the supports' resulting embeddings using a pre-defined distance metric.\\
As an extension for the few-shot case, Snell et al. introduced \textbf{prototypical networks} in 2017 \cite{snell}. Instead of treating all of a class' support examples independently, they compute a ``prototype'' of the class after the embedding network, which is usually the centroid of the support embeddings.\\
When applying few-shot learning to a binary problem, as in our case, support examples for two classes are required: Positive examples of the class of interest, and random other examples. This amounts to a workaround, as there are no two comparable classes. Instead, one of them merely consists of counterexamples (a sort of ``garbage'' class). Examples in this class will have a much wider feature distribution. This makes selecting random examples difficult, as they should still be representative of this distribution, particularly when only a small number of support examples is used. For this reason, we test an extension to prototypical models, in which the positive class is modeled as described, but the negative class is only represented by a non-trainable centroid at the origin. Episodes in this training mode do not need to contain negative support examples. We call this architecture \textbf{one-way prototypical networks}.\\
In all cases, we modify the original architecture to be applicable to short-text classification. We do this by replacing the proposed embedding networks in all branches with Kim's CNN \cite{kim}, which is used frequently in related tasks as described in section \ref{sec:sota}. All parameters are set as described in the original publication. The original softmax output is then replaced with the appropriate distance metric (matching/prototypical/one-way prototypical).\\
Infinite Mixture Prototypes \cite{allen} were also tested, but did not lead to good results. There is, however, some evidence that replacing prototype centroids with Gaussian distributions can produce higher scores \cite{iscram19}. We are currently investigating this more closely.

\section{Experimental results}
\vspace{-7pt}
As described in section \ref{subsec:data_comp}, we perform experiments on two training configurations: Keyword-based detection of tweets concerning a certain event vs. tweets concerning other events, and detection of tweets concerning a certain event vs. random other tweets. Training is done on a set of 12,800 randomly generated episodes over 20 epochs, and validation is performed on 6,400 episodes. All experiments are performed with five different initializations, results are averaged. Standard deviation between runs is generally very low ($\leq$.01).\\

\begin{figure}[H]
    \centering
    \begin{subfigure}[b]{0.49\textwidth}
        \includegraphics[width=\textwidth]{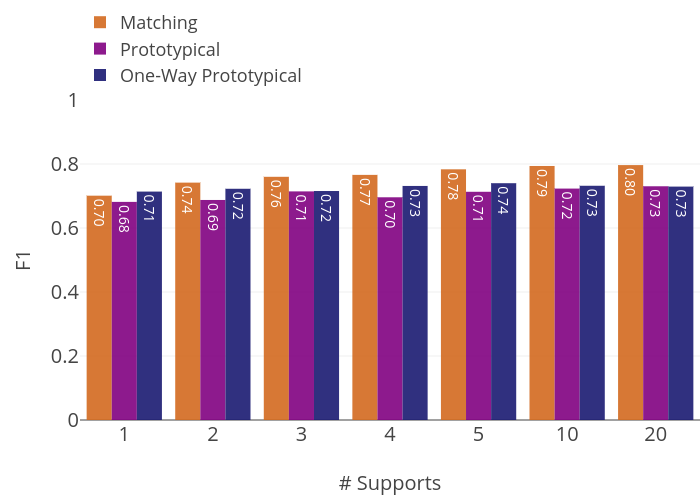}
    \caption{Event-vs.-event configuration}\label{fig:res_kw}
    \end{subfigure}
    \begin{subfigure}[b]{0.49\textwidth}
        \includegraphics[width=\textwidth]{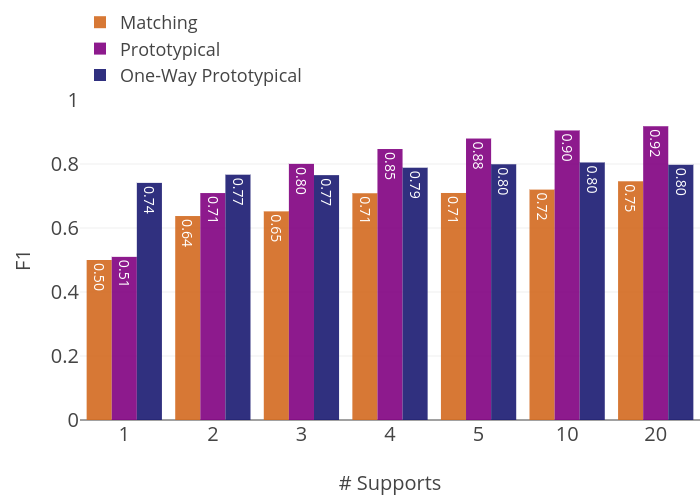}
           \caption{Event-vs.-all configuration}\label{fig:res_ev}
    \end{subfigure}

    \caption{$F_1$ results; matching, prototypical, and one-way prototypical networks.}
\end{figure}

\subsection{Event-vs.-event training}
Figure \ref{fig:res_kw} shows the $F_1$ results for the event-vs.-event classification task using matching, prototypical, and one-way prototypical networks. The difficulty of this task lies in the fact that the positive and negative class are highly related, and may frequently cover similar topics, only for different events. On the other hand, the classes have a large internal spread as crisis events spawn many different sub-topics. In conclusion, the overlap between both classes is large, and classifying tweets can even be difficult for humans.\\
Interestingly, prototypical networks perform worse for this task than matching networks; this is usually not the case in other experiments. It makes sense here, though, because the two class prototypes may end up very close together due to the mentioned overlap. The nearest-neighbor mechanic of matching networks then works better. One-way prototypical networks perform similarly well when few support examples are available. They do not share the overlap problem because the negative class is fixed at the origin. (Side note: The difference between matching and prototypical networks for one support example arises because they use Cosine and Euclidean distance respectively).

\subsection{Event-vs.-all training}
The results for our second experiment, the detection of event-related tweets in random other tweets, are shown in figure \ref{fig:res_ev}. The problem here is positioned somewhat differently, as the positive and negative class should not overlap very much, and the negative class has a much wider distribution. Consequently, prototypical networks perform better here than matching networks because they can generate more distinctive centroids of the positive class here, whereas the nearest-neighbor approach in matching networks fails for the wide negative class. For one support example, neither of them performs better than chance due to the fact that the negative support example could represent anything. In this case, the one-way prototypical networks generate useful results. Beyond three support examples, they are overtaken by two-way prototypical networks; it seems to be possible to triangulate a useful negative centroid at this point.\\
This appears to be the more realistic application, as there are rarely multiple events happening in parallel, and we are interested in detecting event-related tweets out of a multitude of other tweets. With just 10 support examples, we achieve an $F_1$ measure of $.9$, which might be higher in a real-world scenario where the examples are hand-picked rather than random. This is also higher than the reported state-of-the-art, albeit on a somewhat differently posed problem.

\section{Conclusion \& future work}
In this paper, we consider the problem of detecting tweets pertaining to a specific disaster based on a small number of example tweets. Research so far has mainly focused on building models to detect any disaster-related tweet instead of those for particular event; however, this can produce much more relevant results than a general-purpose model. In emerging events, a large amount of training data may not be available yet and collecting it may be too time-consuming. We therefore test few-shot models for this task. As an additional difficulty, we are only interested in a single class; its negative counterpart needs to be defined with random other examples, or implicitly covered by the model. Three few-shot approaches are evaluated: Matching networks, prototypical networks, and one-way prototypical networks.\\
Experiments are performed on tweets from one event vs. from other events, and on tweets for one event vs. random other tweets. In the first scenario, prototypical networks perform worse than matching networks because the prototypes are too close. In the second, more realistic set-up, they perform better and deliver over-all higher results, beating the state-of-the-art for detecting disaster tweets. One-way protoypical networks work particularly well in both cases when very few support examples are available.\\
In the future, these methods could be improved in a number of ways. Ideally, they could be trained on a much wider range of (disaster) events as this would help their generalization abilities towards unseen events. When possible, the negative examples could be picked in a more directed fashion to stabilize the networks' decision boundaries, or the network could simply be enabled to process a much larger amount of negative examples than positive examples without changing the class priors. Finally, we are currently analyzing the one-way prototypical method more closely for other tasks.

\newpage
\bibliographystyle{abbrv}
\bibliography{references}

\begin{thebibliography}{10}

\bibitem{alam}
F.~Alam, S.~Joty, and M.~Imran.
\newblock Domain adaptation with adversarial training and graph embeddings.
\newblock In {\em 56th Annual Meeting of the Association for Computational
  Linguistics (ACL)}, Melbourne, Australia, July 2018.

\bibitem{allen}
K.~R. Allen, E.~Shelhamer, H.~Shin, and J.~B. Tenenbaum.
\newblock Infinite mixture prototypes for few-shot learning.
\newblock In {\em International Conference on Machine Learning (ICML)}, Long
  Beach, CA, USA, June 2019.

\bibitem{redcross}
{American Red Cross}.
\newblock Social media in disasters and emergencies, Aug. 2010.

\bibitem{burel}
G.~Burel and H.~Alani.
\newblock {Crisis Event Extraction Service (CREES) - Automatic Detection and
  Classification of Crisis-related Content on Social Media}.
\newblock In {\em 15th International Conference on Information Systems for
  Crisis Response and Management (ISCRAM)}, Rochester, NY, USA, May 2018.

\bibitem{burel2}
G.~Burel, H.~Saif, and H.~Alani.
\newblock Semantic wide and deep learning for detecting crisis-information
  categories on social media.
\newblock In {\em International Semantic Web Conference {(ISWC)}}, Vienna,
  Austria, Oct. 2017.

\bibitem{caragea}
C.~Caragea, A.~Silvescu, and A.~H. Tapia.
\newblock Identifying informative messages in disaster events using
  convolutional neural networks.
\newblock In {\em 13th International Conference on Information Systems for
  Crisis Response and Management (ISCRAM)}, Rio de Janeiro, Brazil, May 2016.

\bibitem{chen}
Y.~Chen, L.~Xu, K.~Liu, D.~Zeng, and J.~Zhao.
\newblock Event extraction via dynamic multi-pooling convolutional neural
  networks.
\newblock In {\em Annual Meeting of the Association for Computational
  Linguistics (ACL)}, July 2015.

\bibitem{feng}
X.~Feng, B.~Qin, and T.~Liu.
\newblock A language-independent neural network for event detection.
\newblock In {\em 54th Annual Meeting of the Association for Computational
  Linguistics (ACL)}, Berlin, Germany, Aug 2016.

\bibitem{go}
A.~Go, R.~Bhayani, and L.~Huang.
\newblock Twitter sentiment classification using distant supervision.
\newblock Technical report, Stanford University, 2009.

\bibitem{imran2}
M.~Imran, C.~Castillo, F.~Diaz, and S.~Vieweg.
\newblock Processing {Social} {Media} {Messages} in {Mass} {Emergency}: {A}
  {Survey}.
\newblock {\em ACM Computing Surveys}, 47(4):1--38, June 2015.

\bibitem{imran}
M.~Imran, P.~Mitra, and C.~Castillo.
\newblock Twitter as a lifeline: Human-annotated twitter corpora for nlp of
  crisis-related messages.
\newblock In {\em Tenth International Conference on Language Resources and
  Evaluation (LREC)}, Portoroz, Slovenia, May 2016.

\bibitem{kim}
Y.~Kim.
\newblock Convolutional neural networks for sentence classification.
\newblock In {\em 2014 Conference on Empirical Methods in Natural Language
  Processing (EMNLP)}, Doha, Qatar, Oct. 2014.

\bibitem{iscram19}
A.~Kruspe, J.~Kersten, and F.~Klan.
\newblock Detecting event-related tweets by example using few-shot models.
\newblock In {\em International Conference on Information Systems for Crisis
  Response and Management (ISCRAM)}, Valencia, Spain, May 2019.

\bibitem{kumar}
S.~Kumar, G.~Barbier, M.~A. Abbasi, and H.~Liu.
\newblock Tweettracker: An analysis tool for humanitarian and disaster relief.
\newblock In {\em International AAAI Conference on Weblogs and Social Media
  (ICWSM)}, Barcelona, Spain, June 2011.

\bibitem{landwehr}
P.~M. Landwehr and K.~M. Carley.
\newblock Social {Media} in {Disaster} {Relief} - usage patterns, data mining
  tools, and current research directions.
\newblock In W.~W. Chu, editor, {\em Data {Mining} and {Knowledge} {Discovery}
  for {Big} {Data}}, pages 225--257. Springer Berlin Heidelberg, Berlin,
  Heidelberg, Jan. 2014.

\bibitem{mcminn}
A.~J. McMinn, Y.~Moshfeghi, and J.~M. Jose.
\newblock Building a large-scale corpus for evaluating event detection on
  twitter.
\newblock In {\em ACM International Conference on Information and Knowledge
  Management (CIKM)}, pages 409--418, San Francisco, CA, USA, Oct. 2013.

\bibitem{nguyen}
D.~T. Nguyen, S.~R. Joty, M.~Imran, H.~Sajjad, and P.~Mitra.
\newblock Applications of online deep learning for crisis response using social
  media information.
\newblock In {\em International Workshop on Social Web for Disaster Management
  (SWDM)}, Los Angeles, CA, USA, Oct. 2016.

\bibitem{nguyen2}
T.~H. Nguyen and R.~Grishman.
\newblock Event detection and domain adaptation with convolutional neural
  networks.
\newblock In {\em 53rd Annual Meeting of the Association for Computational
  Linguistics (ACL)}, Beijing, China, July 2015.

\bibitem{niles}
M.~T. Niles, B.~F. Emery, A.~J. Reagan, P.~S. Dodds, and C.~M. Danforth.
\newblock Social media usage patterns during natural hazards.
\newblock {\em PLOS ONE}, 14(2):1--16, 02 2019.

\bibitem{olteanu}
A.~Olteanu, C.~Castillo, F.~Diaz, and S.~Vieweg.
\newblock Crisislex: A lexicon for collecting and filtering microblogged
  communications in crises.
\newblock In {\em AAAI Conference on Weblogs and Social Media (ICWSM)}, Ann
  Arbor, MI, USA, June 2014.

\bibitem{olteanu2}
A.~Olteanu, S.~Vieweg, and C.~Castillo.
\newblock What to expect when the unexpected happens: Social media
  communications across crises.
\newblock In {\em Conference on Computer Supported Cooperative Work and Social
  Computing (ACM CSCW)}, Vancouver, BC, Canada, Mar. 2015.

\bibitem{rogstadius}
J.~Rogstadius, M.~Vukovic, C.~A. Teixeira, V.~Kostakos, E.~Karapanos, and J.~A.
  Laredo.
\newblock {CrisisTracker}: {Crowdsourced} social media curation for disaster
  awareness.
\newblock {\em IBM Journal of Research and Development}, 57(5):4:1--4:13, Sept.
  2013.

\bibitem{sloan}
L.~Sloan, J.~Morgan, W.~Housley, M.~Williams, A.~Edwards, P.~Burnap, and
  O.~Rana.
\newblock Knowing the tweeters: Deriving sociologically relevant demographics
  from twitter.
\newblock {\em Sociological Research Online}, 18(3):7, Aug. 2013.

\bibitem{snell}
J.~Snell, K.~Swersky, and R.~S. Zemel.
\newblock Prototypical networks for few-shot learning.
\newblock In {\em International Conference on Neural Information Processing
  Systems (NIPS)}, Long Beach, CA, USA, Dec. 2017.

\bibitem{vinyals}
O.~Vinyals, C.~Blundell, T.~Lillicrap, K.~Kavukcuoglu, and D.~Wierstra.
\newblock Matching networks for one shot learning.
\newblock In {\em International Conference on Neural Information Processing
  Systems (NIPS)}, Barcelona, Spain, Dec. 2016.

\end{thebibliography}

\newpage
\section*{Appendix A: Event-specific hashtags chosen for the \textit{CrisisLexT26} and \textit{CrisisLex} datasets}
\setlength{\abovecaptionskip}{100pt}

\begin{table}[H]
  \centering
\begin{tabular}{ | l | l | l |} \hline
  \textbf{Event} & \textbf{Hashtag} & \textbf{Occurrences/1001} \\ \hline
  2012 Typhoon Pablo & \#PabloPH & 453 \\
  2013 Bohol Earthquake & \#PrayForVisayas & 338 \\
  2013 Singapore Haze & \#sghaze & 667 \\
  2013 West Texas Explosion & \#PrayForTexas & 152 \\
  2012 Italy Earthquakes & \#terremoto & 711 \\
  2013 Manila Floods & \#MaringPH & 399 \\
  2013 Boston Bombings & \#PrayForBoston & 259 \\
  2013 Brazil Nightclub Fire & \#SantaMaria & 353 \\
  2013 Colorado Floods & \#coflood & 317 \\
  2013 LA Airport Shootings & \#LAX & 451 \\
  2012 Guatemala Earthquake & \#sismo & 165 \\
  2012 Philippines Floods & \#rescuePH & 571 \\
  2013 Sardinia Floods & \#Sardegna & 764 \\
  2012 Venezuela Refinery Explosion & \#Amuay & 588 \\
  2013 Alberta Floods & \#yycflood & 497 \\
  2013 Lac Megantic Train Crash & \#LacMegantic & 254 \\
  2013 Typhoon Yolanda & \#Haiyan & 264 \\
  2013 Glasgow Helicopter Crash & \#Clutha & 286 \\
  2013 Queensland Floods & \#bigwet & 684 \\
  2012 Colorado Wildfires & \#colorado & 151 \\
  2013 Australia Bushfire & \#nswfires & 481 \\
  2013 Savar Building Collapse & \#Bangladesh & 579 \\
  2012 Costa Rica Earthquake & \#earthquake & 363 \\
  2013 Russia Meteor & \#RussianMeteor & 407 \\ \hline
\end{tabular}
\caption{Hashtags chosen for the \textit{CrisisLexT26} data set. (Two events did not contain frequent hashtags and were therefore excluded.)} \label{tab:hashtags_crisislex}
\end{table}

\begin{table}[H]
  \centering
\begin{tabular}{ | l | l | l |} \hline
  \textbf{Event} & \textbf{Hashtag} & \textbf{Occurrences/Total} \\ \hline
  2013 Pakistan Earthquake & \#Balochistan & 402/2015 \\
  2014 California Earthquake & \#napa & 272/2016 \\
  2014 Chile Earthquake & \#PrayForChile & 452/2014 \\ 
  2014 Ebola Virus & \#ebola & 999/2018 \\
  2014 Hurricane Odile & \#Odile & 440/2016 \\
  2014 India Floods & \#india & 250/2009 \\
  2014 MERS & \#MERS & 589/2021 \\
  2014 Pakistan Floods & \#KashmirFloods & 218/2016 \\
  2014 Typhoon Hagupit & \#hagupit & 439/2016 \\
  2015 Cyclone Pam & \#vanuatu & 336/2014 \\
  2015 Nepal Earthquake & \#NepalQuake & 309/3022 \\
  \hline
\end{tabular}
\caption{Hashtags chosen for the \textit{CrisisNLP} data set.} \label{tab:hashtags_crisisnlp}
\end{table}

\end{document}